\documentclass[11pt,a4paper]{article}
\usepackage[hyperref]{acl2021}
\usepackage{times}
\usepackage{latexsym}
\usepackage{comment}
\usepackage{multirow}
\usepackage{booktabs}
\usepackage{graphicx}
\usepackage{fdsymbol}
\usepackage{subfig}

\usepackage{microtype}

\aclfinalcopy 

\newcommand\blfootnote[1]{%
  \begingroup
  \renewcommand\thefootnote{}\footnote{#1}%
  \addtocounter{footnote}{-1}%
  \endgroup
}

\title{Low-Resource Adaptation of Open-Domain Generative Chatbots}

\author{Greyson Gerhard-Young\textsuperscript{$\medstar\clubsuit\varheartsuit$}, Raviteja Anantha\textsuperscript{$\medstar\vardiamondsuit$}, Srinivas Chappidi\textsuperscript{$\vardiamondsuit$}, Björn Hoffmeister\textsuperscript{$\spadesuit\varheartsuit$}\\
\textsuperscript{$\vardiamondsuit$}{Apple}\\
\textsuperscript{$\clubsuit$}{Brown University}\\
\textsuperscript{$\spadesuit$}{Amazon}\\
}

\date{}

\begin{document}
\maketitle
\begin{abstract}
Recent work building open-domain chatbots has demonstrated that increasing model size improves performance~\cite{meena-2020, blender-2020}. 
On the other hand, latency and connectivity considerations dictate the move of digital assistants on the device~\cite{ios15-2021}. Giving a digital assistant like Siri, Alexa, or Google Assistant the ability to discuss just about anything leads to the need for reducing the chatbot model size such that it fits on the user’s device. We demonstrate that low parameter models can simultaneously retain their general knowledge conversational abilities while improving in a specific domain.
Additionally, we propose a generic framework that accounts for variety in question types, tracks reference throughout multi-turn conversations, and removes inconsistent and potentially toxic responses. Our framework seamlessly transitions between chatting and performing transactional tasks, which will ultimately make interactions with digital assistants more human-like.
We evaluate our framework on 1 internal and 4 public benchmark datasets using both automatic (Perplexity) and human (SSA -- Sensibleness and Specificity Average) evaluation metrics and establish comparable performance while reducing model parameters by 90\%.

\end{abstract}

\blfootnote{$\medstar$ Equal contribution.}
\blfootnote{$\varheartsuit$ Work done while at Apple.}

\section{Introduction}
\begin{figure}[t!]
  \includegraphics[width=\columnwidth]{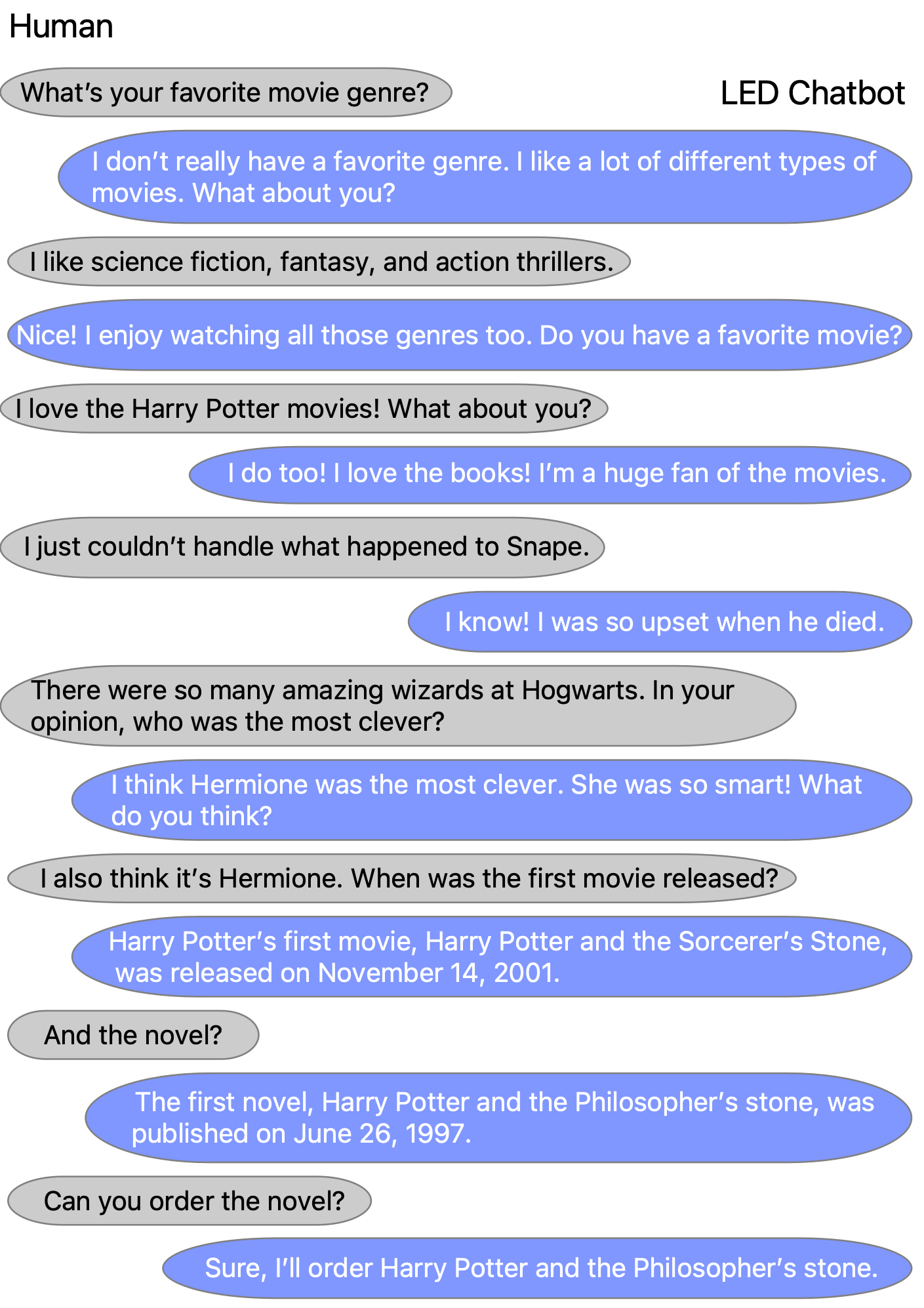}
  \caption{A sample dialogue of paper author (left) conversing with our LED chatbot framework (right). The responses are from the pipeline of models: Reference Resolution, Factual Classifier, Subjective Response Generator, ExtractNParaphrase, Inconsistency/Toxicity Module.}
  \label{fig:sampleconv}
\vspace*{-0.25in}
\end{figure}
Recent progress on end-to-end neural approaches for building open-domain chatbots~\cite{dialo-2020, meena-2020, blender-2020} has demonstrated that large-scale pre-training using heavy-weight models combined with careful selection of datasets for fine-tuning to acquire specific skills can deliver superior performance. However, for one model to perform several tasks --- such as dialogue state tracking or reference resolution, response generation, mitigating toxic responses, avoiding in-turn contradictions, and avoiding incorrect or ``I don't know" responses due to lack of knowledge --- in a reliable fashion, there is still a long way to go. Despite much research, these limitations from the recently proposed approaches prevent practical adoption. In addition, due to huge model sizes, these approaches lack practical utility in a low-resource setting. 

\begin{figure*}[t!]
 \centering
  \includegraphics[scale=0.1]{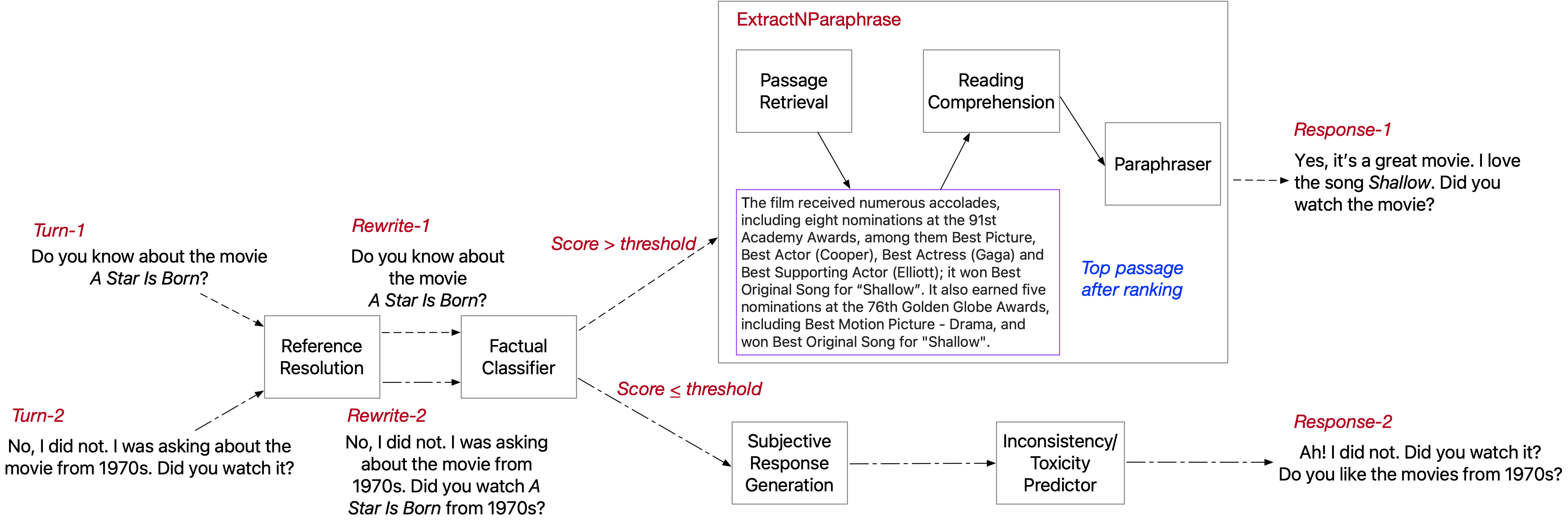}
  \caption{LED Pipeline illustrating end-to-end processing of multi-turn requests and response generation.}
  \label{fig:LED-pipeline}
\vspace*{-0.2in}
\end{figure*}

Some complex frameworks~\cite{milabot-2017, mitsuku-2018, xiaoice-2019} use a mix of templates and dialogue managers with rule-based systems. These complex frameworks often have problems: the produced responses are vague and generic, and they lack engagingness~\cite{meena-2020}. Other complex frameworks address this issue by employing modularizing design assigning each conversational task to a specific component, which can help improve overall performance of the dialogue systems~\cite{uwash-2017, gunrock-2019}. Prior works have shown that generative neural response models outperform template-based or hybrid response generation methods as measured using various human evaluation techniques~\cite{meena-2020, blender-2020}.

In this work, we propose a generic, modular and light-weight framework that blends the desired characteristics of both classes of methods. A snippet of sample dialogue with our proposed framework is shown in Figure~\ref{fig:sampleconv}. Our contributions are as follows: (1) demonstrating that a light-weight response generation model in a modular framework achieves comparable performance to recent models~\cite{meena-2020, blender-2020} that have billions of parameters; (2) providing evidence that adding a reference resolution component improves the quality of the generated response for multi-turn conversations, compared to previous approaches that state track conversational context explicitly or use latent representations~\cite{nlgcontext-2019, blender-2020}; (3) providing a generic end-to-end framework that can process both objective (factual) and subjective questions. 

\section{Lightweight Entertainment Domain Chatbot}
\label{section:LED}
Lightweight Entertainment Domain (LED) chatbot interacts with the user through a pipeline of models. The LED chatbot architecture is illustrated in Figure~\ref{fig:LED-pipeline}. Each module in our pipeline architecture handles a specific conversational task and passes the output for further processing to the downstream modules. In the following subsections, we describe these modules with their respective tasks and training details.

\subsection{Reference Resolution}
In a multi-turn dialogue, the follow-up questions often contain implicit or explicit references to the entities from the previous turns. It is well established that providing self-contained questions by resolving references improves the efficiency of the language understanding systems~\cite{canard-2019, qrecc-2021}.
\begin{figure}[h]
  \includegraphics[width=\columnwidth]{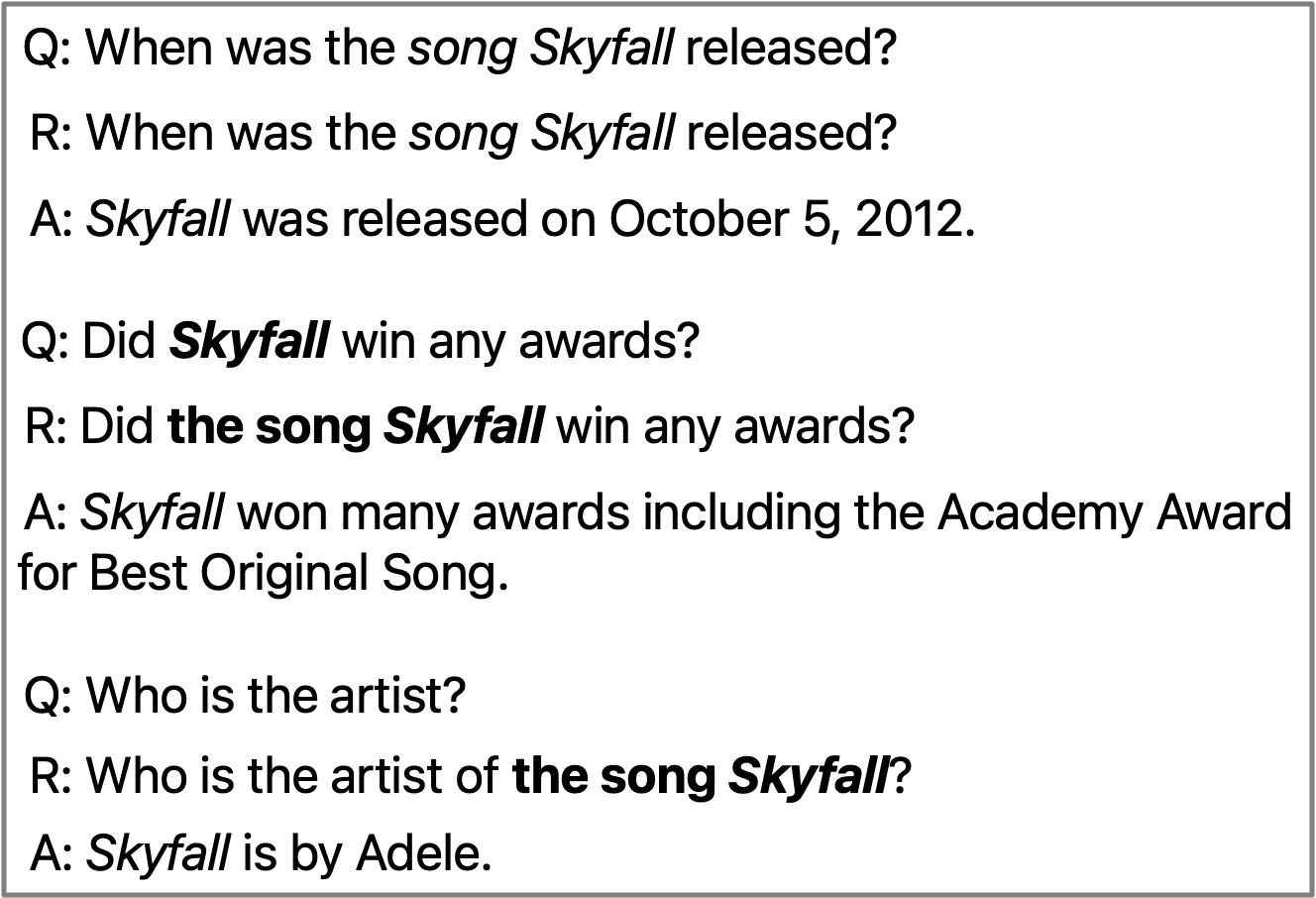}
  \caption{A illustration of reference resolution where the entity reference (in \textbf{bold}) in the question (Q) is disambiguated (Skyfall song vs Skyfall movie) by adding the entity type (song). The rewritten question (R) is a self-contained version of the follow-up question, that will be used for answering (A), where both the co-references and ellipses (in \textbf{bold}) are resolved.}
  \label{fig:refresolveconv}
\vspace*{-0.1in}
\end{figure}

The input to the reference resolution component is the current turn query along with the conversation context, i.e., previous queries and responses. We follow the implementation of the CopyTransformer model~\cite{qrecc-2021}. Our reference resolution model consists of 90M parameters. A sample of input and output is shown in Figure~\ref{fig:refresolveconv}.

\subsection{Factual Classifier}
One of the goals in a low-latency setting is to process a maximum amount of information on the device, and only send to server if it is absolutely needed. This design approach provides faster responses by avoiding unnecessary round trips to the server. In order to determine if the query can be processed on the device it is important to predict if the query needs information from external knowledge sources, such as the world wide web. We refer to the questions that require general knowledge and are of type objective as ``Factual Questions," and the questions that are of type chit-chat as ``Subjective Questions." We refer to the on-device classifier that predicts if a question is factual or not (subjective) as ``Factual Classifier".

We use ALBERT~\cite{albert-2020} as our factual classifier. We initialize the factual classifier weights using HuggingFace pre-trained ALBERT\footnote{\url{https://huggingface.co/albert-base-v2}} model and train using binary labels from our Internal Media dataset, where 1 represents a factual question and 0 a subjective question. Our factual classifier consists of 11M parameters. We observed the optimal value for the threshold to be 0.8.

\subsection{Subjective Response Generation}
The subjective response generation component of our pipeline is a 90M parameter model with a conventional Seq2Seq Transformer architecture. Our work uses the optimized setup discussed in Blender to convert input sequences of dialogue to an output response~\cite{ blender-2020}. However, there are a couple core differences. Our dialogue model was fine-tuned for a particular use case: subjective entertainment-domain questions. Additionally, our model has been trained on rewritten inputs (given our reference resolver in a prior portion of the pipeline). 

The core response generation model was trained using the ParlAI \footnote{\url{https://github.com/facebookresearch/ParlAI}} framework, a platform designed specifically for dialogue models. We build upon the work of Blender’s 90M generative model included in the broader ParlAI zoo~\cite{ blender-2020}. The critical objective for this portion of the pipeline was to maintain general-domain performance while concurrently improving in our target domains: music and movies. As described in Section~\ref{section:data}, our datasets contain human rewritten questions where anaphoric references are resolved, and we use the rewritten questions as input for the response generation. 


\begin{figure}[h]
  \includegraphics[width=\columnwidth]{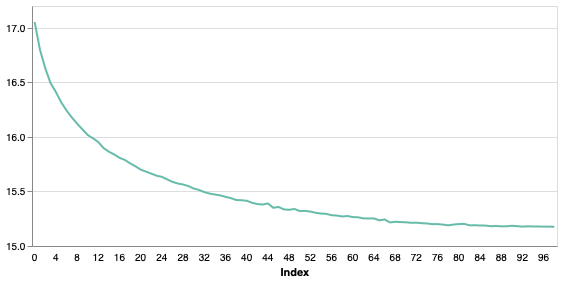}
  \caption {Validation perplexity of subjective response generation model using all five datasets: Wizard of Wiki, ConvAI2, Empathetic Dialogues, Blended Skill Talk, and our internal media dataset with rewritten questions as input.}
  \label{fig:perplexity}
\vspace*{-0.25in}
\end{figure}

\vspace{\baselineskip}
Our experimentation uses a variety of different techniques, with the methodology behind each tactic covered in this section. In order to understand how our fine-tuned model performed on both explicit and implicit inputs, we run all trials on original and rewritten questions (before comparing performance). The tests draw upon common tactics in transfer learning and dialogue models: comparisons on freezing different numbers of layers, retaining the original datasets, and selecting a decoding algorithm. 


\begin{figure}[h]
  \includegraphics[width=\columnwidth]{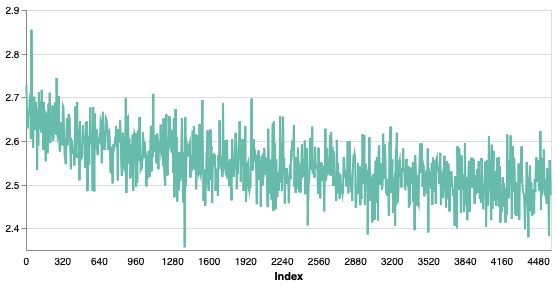}
  \caption {Validation loss of subjective response generation model using all five datasets: Wizard of Wiki, ConvAI2, Empathetic Dialogues, Blended Skill Talk, and our internal media dataset with rewritten questions as input.}
  \label{fig:loss}
\vspace*{-0.25in}
\end{figure}

\vspace{\baselineskip}
In all experiments, we freeze the encoder portion of Blender’s architecture to maintain their well-tuned representation. We compare results between training on the entire decoder and locking its first four layers. In separate automatic evaluation, we contrast using only internal media data to simply adding it as a fifth dataset. Finally, we look at the relative effect of the beam search and Top-K decoding algorithms on human evaluation. The validation perplexity and loss curves of the best run are shown in Figures~\ref{fig:perplexity} and \ref{fig:loss} respectively.

\subsection{ExtractNParaphrase}
In principle, any generative response module is bound to fail when a knowledge-based question is presented and if the response module does not have access to factual information. In our architecture, we route factual questions to the ExtractNParaphrase module, which extracts the answer spans and paraphrases the relevant text to generate a natural and engaging response. The response path for Turn-1 in Figure~\ref{fig:LED-pipeline} illustrates the processing of the question.

ExtractNParaphrase consists of three stages: (1) Passage Retrieval, (2) Reading Comprehension and (3) Paraphrasing. The first two steps follow \citeauthor{qrecc-2021}; and for the third step, paraphrasing, we take motivation from the refine step of \cite{rnr-2018}. We use BM25 to retrieve Top-K passages and a light-weight BERT-based model to extract answer spans. The scores obtained from passage retrieval and answer span extraction are combined to produce the final score. Passage retrieval and answer extraction models are comprised of 50M parameters. We refer to \cite{qrecc-2021} for more details. Finally, we train a sentence paraphraser model based on Transformer, which is comprised of 24M parameters. The paraphrased labels are provided as part of internal media dataset, which is described in Section~\ref{section:data}.

\subsection{Inconsistency/Toxicity Predictor}
Logical consistency in dialogue and avoiding unnecessary or potentially toxic responses are critical factors to consider when developing open-domain chatbots. When interacting with chatbots, people expect coherent responses that at least do not contradict the chatbot’s earlier responses in the same conversation. 

We train a classifier that can detect inconsistent responses given the conversation context. We follow the training procedure described in \cite{nie-2020} using DECODE\footnote{\url{https://parl.ai/projects/contradiction/}} dataset and internal media dataset. We use the ALBERT~\cite{albert-2020} model for inconsistency/toxicity predictor.





\section{Training Data}
\label{section:data}
We use various datasets for training and evaluation focused on different tasks. In this section, we describe each dataset along with the corresponding modules that use the dataset for training.



\textbf{QReCC}~\cite{qrecc-2021} contains around 81,000 conversation turns. Every turn contains a question which may have anaphoric references, a rewritten version of the question with references resolved, an answer span to the question and a corresponding web URL. QReCC data is used to train the reference resolution, passage retrieval and answer span extraction models.

\textbf{Wizard of Wikipedia}~\cite{wow-2019} (WoW) contains 194,000 turns of dialogue distributed over 1,250 topics. Each conversation is predicated on discussing the relevant topic in depth, with the goal of displaying expert knowledge in that subject. Note that in our pipeline framework, we refer objective questions to the ExtractNParaphrase component, so the subjective response generation model is not required to answer factual questions with a high degree of accuracy. Still, the WoW dataset helps our generative model maintain a breadth of knowledge to provide pertinent answers to subjective inputs.

\textbf{ConvAI2} is based off of the work of PersonaChat~\cite{convai-1-2018, convai2-2019} and was used at the NeurIPS 2018 ConvAI competition. This dataset is made up of 140,000 turns where gatherers are given a persona and tasked with learning about their counterpart. This helps open-domain agents ask questions, and perhaps more relevantly in our use case, respond in an engaging manner. We use the ConvAI2 dataset to train the subjective response generation model.

\textbf{Empathetic Dialogues}~\cite{ed-2019} (ED) is a library of 50,000 turns where one speaker plays the role of sympathetic listener. These skills translate well to our needs, as the subjective model must account for previous dialogue history and attempt to match their chosen response to the appropriate tone.

\textbf{Blended Skill Talk}~\cite{bst-2020} (BST) is a 76,000 turn compilation of the previous three datasets: WoW, ConvAI2, and ED. Guided human speakers were given the option to select between outputs from models trained on each of the individual tasks, which produces data that can teach the bot when a certain class of response should be used.

\textbf{DECODE}~\cite{nie-2020} is a conversational dataset made up of 40,280 turns from human to human and human to bot contradictory dialogues. We use DECODE to train the inconsistency/toxicity detector model based off of the ALBERT model, along with our internal media dataset.  

\textbf{{Internal Media dataset}} is composed of 100,000 movie themed turns. Each turn contains a natural question without explicit reference to the movie being discussed, as well as rewritten questions that convert those references to specifics (akin to the reference resolution component of our pipeline). Answer span along with web URL as well as paraphrased variation that is natural and engaging is also provided. 

The dataset is collected using crowd-sourced annotators. The goal of the annotators is to mimic the flow of a natural human conversation while maintaining a neutral persona. The responses were validated against guidelines to be non-controversial, eliminate profanity, be neutral, engaging and concise (with an upper bound of 30 words). Every conversation consists of 10 turns, and we collect 10,000 conversations. We give instructions to explicitly add anaphoric references in follow up turns.  

\section{Evaluation Metrics and Results}
We categorize our evaluation metrics based on component-wise vs end-to-end evaluation. QReCC and DECODE datasets are only used for task-specific model training and are not used in establishing a chatbot's end-to-end metrics: Perplexity and Sensibleness and Specificity Average (SSA). We establish a human evaluation metric, SSA, on our internal media dataset only, due to limited human annotators. We establish the automatic evaluation metric, perplexity, on all 5 datasets: WoW, ConvAI2, ED, BT, and our internal media dataset. Below we discuss the intrinsic (component-wise) and extrinsic (end-to-end) metrics used to evaluate our LED framework. 

\subsection{Intrinsic Metrics}
Excluding the subjective response generation model, all other components in LED have their own task-specific evaluation metrics. For reference resolution model using query rewriting and paraphraser in ExtractNParaphrase module, we use ROUGE, USE and Recall@10 as described in \cite{qrecc-2021}. For factual classifier and inconsistency/toxicity predictor, we use F1 as the evaluation metric and obtain 0.94 and 0.61 respectively. For passage retrieval of ExtractNParaphrase module we use MRR and Recall@k; similarly for answer-span extraction we use F1 and exact match as described in \cite{qrecc-2021}. For the subjective response generation model we use perplexity, which is also our extrinsic metric.

\subsection{Extrinsic Metrics}
Our chatbot framework uses perplexity as its extrinsic metric for automatic evaluation. While there are a number of evaluation metrics that can serve to measure the quality of responses (see the other components of our pipeline), perplexity correlates well with human judgement~\cite{meena-2020}. We build on the work of Meena~\cite{meena-2020} that proposed SSA, Sensibleness and Specificity Average. We use SSA as another extrinsic metric for human evaluation. \citeauthor{meena-2020} subsequently demonstrated a strong correlation between perplexity and SSA among numerous state-of-the-art chatbots. 

Table~\ref{tab:perplexity_metrics} shows perplexity metrics of Blender models, both 90M and 2.7B parameter models; and LED framework, both with and without reference resolution, across all 5 datasets: 1 internal media dataset and 4 public dataset.

\begin{table}[t]
\caption{Comparison of Perplexity metric across various datasets of Blender and LED chatbot frameworks with different parameter size.}
\label{tab:perplexity_metrics}
\centering
\resizebox{\columnwidth}{!}{
\begin{tabular}{lcccc}
  \toprule
 &  &  & LED without & LED with \\
Dataset/Model & Blender 90M & Blender 2.7B & rewritten & rewritten \\
 & &  & input 186M & input 276M \\
 \toprule
Wizard of Wiki & 17.71 & 11.23 & 10.27 & \textbf{9.75} \\
BST & 14.48 & \textbf{8.12} & 8.79 & \textbf{8.54} \\
ConvAI2 & 11.34 & \textbf{7.76} & 8.72 & \textbf{8.01} \\
ED & 11.81 & \textbf{9.83} & 10.31 & \textbf{9.97}\\
Internal Media Dataset & 33.51 & \textbf{15.62} & 18.49 & 16.44\\ 
\bottomrule
\end{tabular}}
\end{table}

Table~\ref{tab:ssa_metrics} shows SSA metrics of Blender models (both 90M and 2.7B parameter models) and LED framework (both with and without reference resolution) on internal media dataset.

\begin{table}[t]
\caption{Comparison of SSA metric and number of model parameters of Blender and LED chatbot frameworks on internal media dataset.}
\label{tab:ssa_metrics}
\centering
\resizebox{\columnwidth}{!}{
\begin{tabular}{lrrrr}
  \toprule
Model/Metric & Parameters & Sensibleness & Specificity & SSA \\
 \toprule
Blender & 90M & 72.60 & 83.10 & 77.85 \\
Blender & 2.7B & \textbf{80.42} & \textbf{92.70} & \textbf{86.56} \\
LED & 186M & 78.28 & 89.12 & 83.70 \\
LED & 276M & \textbf{80.38} & \textbf{91.95} & \textbf{86.17}\\
\bottomrule
\end{tabular}}
\end{table}



%

%

\section{Related Work}
Our work follows the objective of combining open-domain chatbot and transactional digital assistants. The factual classifier component of LED serves as the gatekeeper between these two categories, sending objective asks through the ExtractNParaphrase model and subjective inputs through our fine-tuned open domain model.  While our work broadly falls under the category of open-domain generative chatbots, because of the variety of models and their corresponding tasks, our work also covers multiple key areas in language understanding with a focus on low-resource adaptation design. Prior works~\cite{dialo-2020, meena-2020, blender-2020} have shown that end-to-end neural approaches, where the responses are produced in a generative fashion, can result in engaging dialogue. However, the resultant models from these approaches are huge -- multiple billions of parameters -- and are not on-device friendly. It has also been shown that end-to-end generative chatbots frequently generate responses with inconsistencies~\cite{meena-2020, blender-2020}. It is obvious that there is need for an additional module that can correct, or at least detect, these inconsistencies. Generalizing this approach where we assign a specific task to a module, modularization can lead to overall improvement in dialogue systems~\cite{uwash-2017, gunrock-2019}. We adopt the modularization approach to open-domain generative chatbot to minimize the total number of parameters while tackling some of the shortcomings in the end-to-end neural approaches. 

Blender~\cite{blender-2020} showed non-trivial improvement in response generation when evaluated using human side-by-side comparison. We adopt the Blender model as a basis for the core response generation functionality in subjective cases. We follow the Blender methodology of experimenting with multiple decoding algorithms for optimal human performance. However we also differ from Blender's approach. Firstly, we place a larger emphasis on model size for better on-device compatibility. Secondly, we account for a wider variety of cases where we use answer extraction and paraphrasing to accurately answer factual questions. And finally, we use the reference resolution component to track dialogue state since it is helpful for multi-turn conversations~\cite{qrecc-2021}, along with providing our fine-tuned model with a wider variety of training data (multi-turn conversations where questions are either rewritten or preserved).

Meena~\cite{meena-2020} proposed a new metric, Sensibleness and Specificity Average (SSA), which captures key elements of a human-like multi-turn conversation. Additionally, they also show perplexity is the best automatic metric that correlates well with human judgement. We borrow SSA to evaluate human performance. It is good for our use case, where the model is required not just to answer logically but should also be rewarded for referencing context from earlier in the conversation. One of the differences between our work and Meena is we do not use Evolved Transformer layers, though that may be basis for future work. One difference of our work compared to both Blender and Meena is we follow a modularized approach, instead of a single parameter-heavy model.




\section{Limitations and Future Work}
\subsection{Limitations}
\label{section:limitations}
Although we reduce the number of parameters by 90\% and achieve comparable performance, we still notice shortcomings which can be possibly mitigated by the inconsistency/toxicity classifier. 

\subsubsection{Consistent Agreement}
\begin{figure}[h]
  \includegraphics[width=\columnwidth]{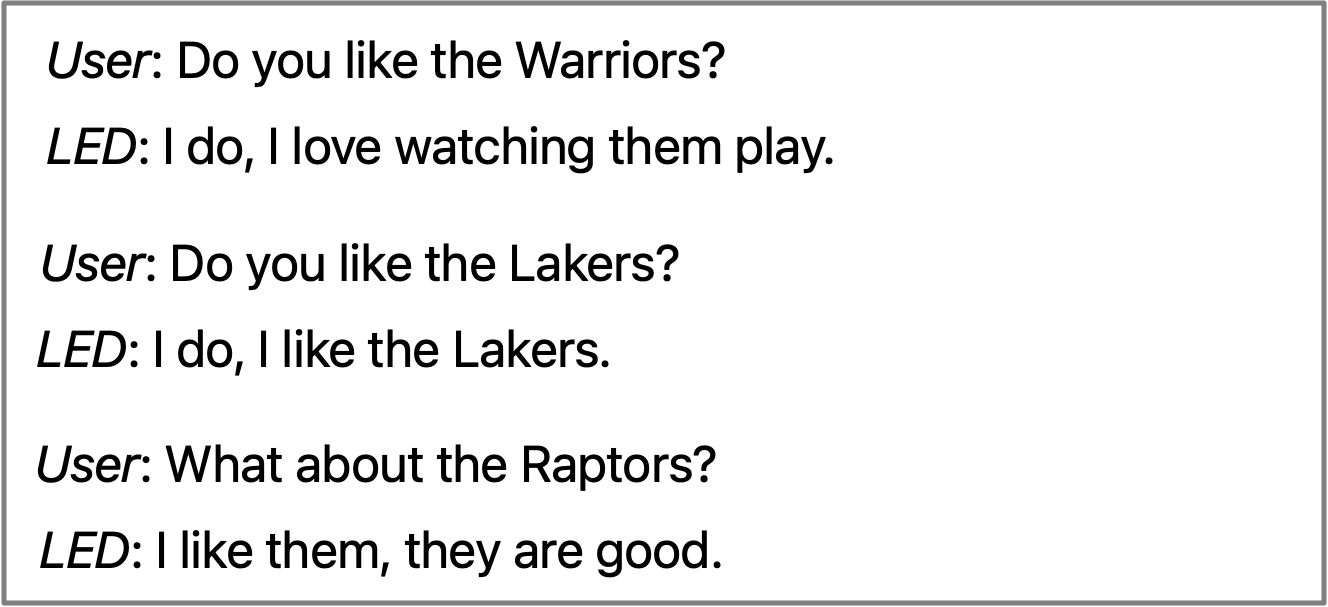}
  \caption {LED in agreement with user the majority of the time.}
  \label{fig:agreement}
\vspace*{-0.25in}
\end{figure}

LED, often, is in agreement with the user which might cause the user to feel non-engaging. This behavior stems from the inclusion of the Empathetic Dialogues~\cite{ed-2019} dataset in the Subjective Response Generation component. Utilized in both the pre-trained Blender model and our fine-tuning process, Empathetic Dialgoues data incentivize the model to choose agreeable responses.  An example of this behavior is shown in Figure~\ref{fig:agreement}.

\subsubsection{Sensitive Issues}
\begin{figure}[h]
  \includegraphics[width=\columnwidth]{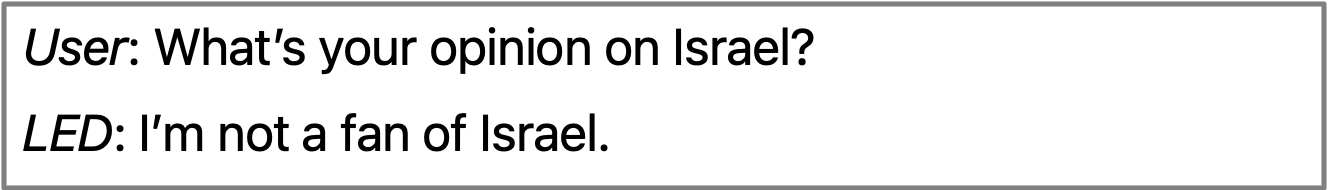}
  \caption {LED responding to controversial question in a non-neutral manner.}
  \label{fig:israel}
\vspace*{-0.25in}
\end{figure}

LED responds to controversial questions with a non-neutral persona. These are instances where the inconsistency/toxicity predictor failed. While this class of responses was frequently present in the Subjective Response Generation component, we were able to significantly mitigate overall prevalence through the inclusion of the inconsistency/toxicity predictor component. An example of such an instance is shown in Figure~\ref{fig:israel}.

\subsubsection{Questionable Advice}
\begin{figure}[h]
  \includegraphics[width=\columnwidth]{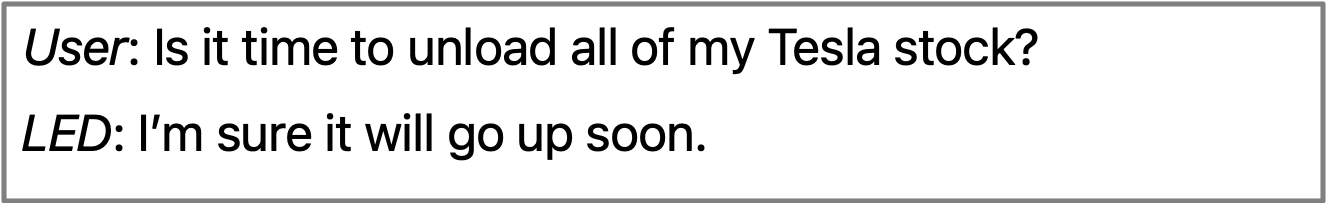}
  \caption {LED providing unnecessary or questionable financial advice.}
  \label{fig:tesla-stock}
\vspace*{-0.1in}
\end{figure}

LED provides unnecessary or questionable advice to questions seeking advice. The root cause of these outputs are examples from the Wizard of Wikipedia~\cite{wow-2019} dataset, where the model is taught to display expert knowledge in a particular area. An example of unnecessary financial advice is shown in Figure~\ref{fig:tesla-stock}. 

\subsection{Future Work}
We plan to investigate solutions to mitigate the undesired patterns noticed in Section~\ref{section:limitations} by improving the inconsistency/toxicity predictor, as well as, investigate the feasibility of a common embedding layer for all modules in our framework in an effort to further minimize the number of parameters with minimum or no-drop in performance.

Also, transactional requests have a stronger user feedback signal (e.g. if playing the wrong movie, then the user will stop the movie), which can help to learn whether a conversation was successful. The conversational models (i.e., natural language understanding) can learn from user feedback signals. We plan to investigate incorporating such feedback signals to improve task completion rate in a conversation.

\section*{Acknowledgements}
We would like to thank Barry Theobald, Russ Webb, Alex Acero and John Giannandrea for their insightful comments.


\bibliographystyle{acl_natbib}
\bibliography{acl2021}

\begin{thebibliography}{20}
\expandafter\ifx\csname natexlab\endcsname\relax\def\natexlab#1{#1}\fi

\bibitem[{Adiwardana et~al.(2020)Adiwardana, Luong, So, Hall, Fiedel,
  Thoppilan, Yang, Kulshreshtha, Nemade, Lu, and Le}]{meena-2020}
Daniel Adiwardana, Minh-Thang Luong, David~R. So, Jamie Hall, Noah Fiedel,
  Romal Thoppilan, Zi~Yang, Apoorv Kulshreshtha, Gaurav Nemade, Yifeng Lu, and
  Quoc~V. Le. 2020.
\newblock \href {https://arxiv.org/pdf/2001.09977.pdf} {Towards a human-like
  open-domain chatbot}.
\newblock arXiv:2001.09977.

\bibitem[{Anantha et~al.(2021)Anantha, Vakulenko, Tu, Longpre, Pulman, and
  Chappidi}]{qrecc-2021}
Raviteja Anantha, Svitlana Vakulenko, Zhucheng Tu, Shayne Longpre, Stephen
  Pulman, and Srinivas Chappidi. 2021.
\newblock Open-domain question answering goes conversational via question
  rewriting.
\newblock In \emph{Proceedings of the 2021 Conference of the North American
  Chapter of the Association for Computational Linguistics: Human Language
  Technologies}, page 520–534.

\bibitem[{Cervone et~al.(2019)Cervone, Khatri, Goel, Hedayatnia, Venkatesh,
  Hakkani-Tür, and Gabriel}]{nlgcontext-2019}
Alessandra Cervone, Chandra Khatri, Rahul Goel, Behnam Hedayatnia, Anu
  Venkatesh, Dilek Hakkani-Tür, and Raefer Gabriel. 2019.
\newblock Natural language generation at scale: A case study for open domain
  question answering.
\newblock In \emph{Proceedings of the 12th International Conference on Natural
  Language Generation}.

\bibitem[{Dinan et~al.(2019{\natexlab{a}})Dinan, Logacheva, Malykh, Miller,
  Shuster, Urbanek, Kiela, Szlam, Serban, Lowe, Prabhumoye, Black, Rudnicky,
  Williams, Pineau, Burtsev, and Weston}]{convai2-2019}
Emily Dinan, Varvara Logacheva, Valentin Malykh, Alexander Miller, Kurt
  Shuster, Jack Urbanek, Douwe Kiela, Arthur Szlam, Iulian Serban, Ryan Lowe,
  Shrimai Prabhumoye, Alan~W Black, Alexander Rudnicky, Jason Williams, Joelle
  Pineau, Mikhail Burtsev, and Jason Weston. 2019{\natexlab{a}}.
\newblock \href {https://arxiv.org/pdf/1902.00098.pdf} {The second
  conversational intelligence challenge (convai2)}.
\newblock arXiv:1902.00098.

\bibitem[{Dinan et~al.(2019{\natexlab{b}})Dinan, Roller, Shuster, Fan, Auli,
  and Weston}]{wow-2019}
Emily Dinan, Stephen Roller, Kurt Shuster, Angela Fan, Michael Auli, and Jason
  Weston. 2019{\natexlab{b}}.
\newblock \href {https://arxiv.org/pdf/1811.01241.pdf} {Wizard of wikipedia:
  Knowledge-powered conversational agents}.
\newblock arXiv:1811.01241.

\bibitem[{Elgohary et~al.(2019)Elgohary, Peskov, and Boyd-Graber}]{canard-2019}
Ahmed Elgohary, Denis Peskov, and Jordan Boyd-Graber. 2019.
\newblock \href {https://www.aclweb.org/anthology/D19-1605.pdf} {Can you unpack
  that? learning to rewrite questions-in-context}.
\newblock In \emph{Proceedings of the 2019 Conference on Empirical Methods in
  Natural Language Processing and the 9th International Joint Conference on
  Natural Language Processing}, pages 5920--5926.

\bibitem[{Fang et~al.(2017)Fang, Cheng, Clark, Holtz-man, Sap, Ostendorf, Choi,
  and Smith}]{uwash-2017}
Hao Fang, Hao Cheng, Elizabeth Clark, Ariel Holtz-man, Maarten Sap, Mari
  Ostendorf, Yejin Choi, and Noah~A Smith. 2017.
\newblock \href
  {https://homes.cs.washington.edu/~msap/pdfs/fang2017alexatechreport.pdf}
  {Sounding board – university of washington’s alexa prize submission}.

\bibitem[{Lan et~al.(2020)Lan, Chen, Goodman, Gimpel, Sharma, and
  Soricut}]{albert-2020}
Zhenzhong Lan, Mingda Chen, Sebastian Goodman, Kevin Gimpel, Piyush Sharma, and
  Radu Soricut. 2020.
\newblock \href {https://arxiv.org/abs/1909.11942} {Albert: A lite bert for
  self-supervised learning of language representations}.
\newblock arXiv:1909.11942.

\bibitem[{Nie et~al.(2020)Nie, Williamson, Bansal, Kiela, and
  Weston}]{nie-2020}
Yixin Nie, Mary Williamson, Mohit Bansal, Douwe Kiela, and Jason Weston. 2020.
\newblock \href {http://arxiv.org/abs/2012.13391} {I like fish, especially
  dolphins: Addressing contradictions in dialogue modelling}.

\bibitem[{Rashkin et~al.(2019)Rashkin, Smith, Li, and Boureau}]{ed-2019}
Hannah Rashkin, Eric~Michael Smith, Margaret Li, and Y-Lan Boureau. 2019.
\newblock Towards empathetic open-domain conversation models: a new benchmark
  and dataset.
\newblock In \emph{Proceedings of the 57th Annual Meeting of the Association
  for Computational Linguistics}, page 5370–5381.

\bibitem[{Roller et~al.(2020)Roller, Dinan, Goyal, Ju, Williamson, Liu, Xu,
  Ott, Shuster, Smith, Boureau, and Weston}]{blender-2020}
Stephen Roller, Emily Dinan, Naman Goyal, Da~Ju, Mary Williamson, Yinhan Liu,
  Jing Xu, Myle Ott, Kurt Shuster, Eric~M. Smith, Y-Lan Boureau, and Jason
  Weston. 2020.
\newblock \href {https://arxiv.org/pdf/2004.13637.pdf} {Recipes for building an
  open-domain chatbot}.
\newblock arXiv:2004.13637.

\bibitem[{Serban et~al.(2017)Serban, Sankar, Germain, Zhang, Lin, Subramanian,
  Kim, Pieper, Chandar, Ke, Rajeshwar, de~Brebisson, Sotelo, Suhubdy,
  Michalski, Nguyen, Pineau, and Bengio}]{milabot-2017}
Iulian~V. Serban, Chinnadhurai Sankar, Mathieu Germain, Saizheng Zhang, Zhouhan
  Lin, Sandeep Subramanian, Taesup Kim, Michael Pieper, Sarath Chandar,
  Nan~Rosemary Ke, Sai Rajeshwar, Alexandre de~Brebisson, Jose M.~R. Sotelo,
  Dendi Suhubdy, Vincent Michalski, Alexandre Nguyen, Joelle Pineau, and Yoshua
  Bengio. 2017.
\newblock \href {https://arxiv.org/abs/1709.02349} {A deep reinforcement
  learning chatbot}.
\newblock arXiv:1709.02349.

\bibitem[{Smith et~al.(2020)Smith, Williamson, Shuster, Weston, and
  Boureau}]{bst-2020}
Eric~Michael Smith, Mary Williamson, Kurt Shuster, Jason Weston, and Y-Lan
  Boureau. 2020.
\newblock Can you put it all together: Evaluating conversational agents’
  ability to blend skills.
\newblock In \emph{Proceedings of the 58th Annual Meeting of the Association
  for Computational Linguistics}, page 2021–2030.

\bibitem[{Verge(2021)}]{ios15-2021}
The Verge. 2021.
\newblock \href
  {https://www.theverge.com/2021/6/7/22522993/apple-siri-on-device-speech-recognition-no-internet-wwdc}
  {Apple’s siri will finally work without an internet connection with
  on-device speech recognition}.

\bibitem[{Weston et~al.(2018)Weston, Dinan, and Miller}]{rnr-2018}
Jason Weston, Emily Dinan, and Alexander~H. Miller. 2018.
\newblock Retrieve and refine: Improved sequence generation models for
  dialogue.
\newblock In \emph{Proceedings of the 2018 EMNLP Workshop SCAI: The 2nd
  International Workshop on Search-Oriented Conversational AI
  978-1-948087-75-9}.

\bibitem[{Worswick(2018)}]{mitsuku-2018}
Steve Worswick. 2018.
\newblock \href
  {https://medium.com/pandorabots-blog/mitsuku-wins-loebner-prize-2018-3e8d98c5f2a7}
  {Mitsuku wins loebner prize 2018!}

\bibitem[{Yu et~al.(2019)Yu, Cohn, Yang, Chen, Wen, Zhang, Zhou, Jesse, Chau,
  Bhowmick, Iyer, Sreenivasulu, Davidson, and andd Zhou~Yu}]{gunrock-2019}
Dian Yu, Michelle Cohn, Yi~Mang Yang, Chun-Yen Chen, Weiming Wen, Jiaping
  Zhang, Mingyang Zhou, Kevin Jesse, Austin Chau, Antara Bhowmick, Shreenath
  Iyer, Giritheja Sreenivasulu, Sam Davidson, and Ashwin~Bhandare andd Zhou~Yu.
  2019.
\newblock Gunrock: A social bot for complex and engaging long conversations.
\newblock In \emph{Proceedings of the 2019 EMNLP and the 9th IJCNLP (System
  Demonstrations)}, page 79–84.

\bibitem[{Zhang et~al.(2018)Zhang, Dinan, Urbanek, Szlam, Kiela, and
  Weston}]{convai-1-2018}
Saizheng Zhang, Emily Dinan, Jack Urbanek, Arthur Szlam, Douwe Kiela, and Jason
  Weston. 2018.
\newblock \href {https://arxiv.org/abs/1801.07243} {Personalizing dialogue
  agents: I have a dog, do you have pets too?}
\newblock arXiv:1801.07243v5.

\bibitem[{Zhang et~al.(2020)Zhang, Sun, Galley, Chen, Brockett, Gao, Gao, Liu,
  and Dolan}]{dialo-2020}
Yizhe Zhang, Siqi Sun, Michel Galley, Yen-Chun Chen, Chris Brockett, Xiang Gao,
  Jianfeng Gao, Jingjing Liu, and Bill Dolan. 2020.
\newblock Dialogpt : Large-scale generative pre-training for conversational
  response generation.
\newblock In \emph{Proceedings of the 58th Annual Meeting of the Association
  for Computational Linguistics: System Demonstrations}.

\bibitem[{Zhou et~al.(2019)Zhou, Gao, Li, and Shum}]{xiaoice-2019}
Li~Zhou, Jianfeng Gao, Di~Li, and Heung-Yeung Shum. 2019.
\newblock \href {https://arxiv.org/pdf/1812.08989.pdf} {The design and
  implementation of xiaoice, an empathetic social chatbot}.
\newblock arXiv:1812.08989.

\end{thebibliography}


\end{document}